\documentclass[runningheads]{llncs}

\usepackage{eccv}

\usepackage{eccvabbrv}
\usepackage{booktabs}
\usepackage{multirow}
\usepackage{makecell}
\usepackage{pifont}
\usepackage{wrapfig}
\usepackage{graphicx}

\usepackage{hyperref}

\begin{document}

\title{Boosting Latent Diffusion Models via Semantic-Disentangled VAE} 

\titlerunning{Boosting Latent Diffusion Models via Semantic-Disentangled VAE}

\titlerunning{Boosting Latent Diffusion Models via Semantic-Disentangled VAE}

\author{John Page$^{*}$ \and
Xuesong Niu$^{*}$ \and
Kai Wu$^{\ddag}$ \and Kun Gai}

\authorrunning{John Page et al.}

\institute{Kolors Team, Kuaishou Technology}

\maketitle

\footnotetext[1]{* These authors contributed equally to this work.}  
\footnotetext[2]{\ddag Corresponding author}       

\begin{abstract}
  Latent Diffusion Models (LDMs) rely heavily on the compressed latent space provided by
Variational Autoencoders (VAEs) for high-quality image generation. Recent studies have
attempted to obtain generation-friendly VAEs by directly adopting alignment strategies
from LDM training, leveraging Vision Foundation Models (VFMs) as representation alignment
targets. However, such alignment paradigms overlook the fundamental differences in
representational requirements between LDMs and VAEs. Simple feature mapping from local
patches to high-dimensional semantics can induce semantic collapse, leading to the loss
of fine-grained attributes. In this paper, we reveal a key insight: unlike LDMs that
benefit from high-level global semantics, a generation-friendly VAE must possess strong
semantic disentanglement capabilities to preserve fine-grained, attribute-level information
in a structured manner. To address this discrepancy, we propose the Semantic-Disentangled
VAE (Send-VAE). Deviating from previous shallow alignment approaches, Send-VAE introduces
a non-linear mapping architecture to effectively bridge the local structures of VAEs and
the dense semantics of VFMs, thereby encouraging emergent disentangled properties in the
latent space without explicit regularization. Extensive experiments establish a new paradigm
for evaluating VAE latent spaces via low-level attribute separability and demonstrate that
Send-VAE achieves state-of-the-art generation quality (FID of 1.21) on ImageNet $256\times256$.
Code is available at \\ https://github.com/Kwai-Kolors/Send-VAE.
  \keywords{Image tokenizer \and Latent diffusion model \and Image synthesis}
\end{abstract}

\begin{figure}[h]
\centering
\small
\begin{minipage}{.49\textwidth}
\includegraphics[width=0.9\linewidth]{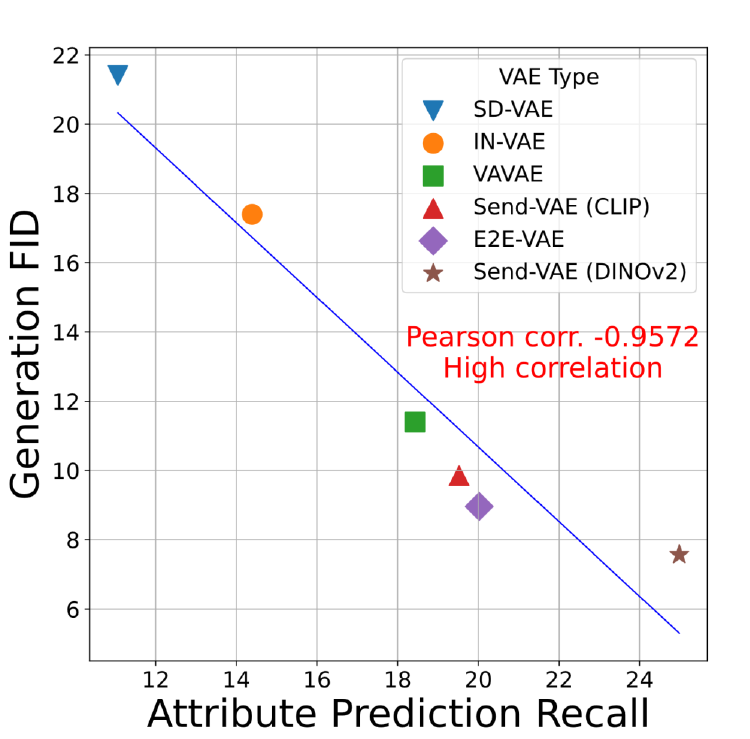}
\setlength{\belowcaptionskip}{-20pt}
\label{fig:method}
\end{minipage}
\ 
\begin{minipage}{.49\textwidth}
\centering
\includegraphics[width=1.\linewidth]{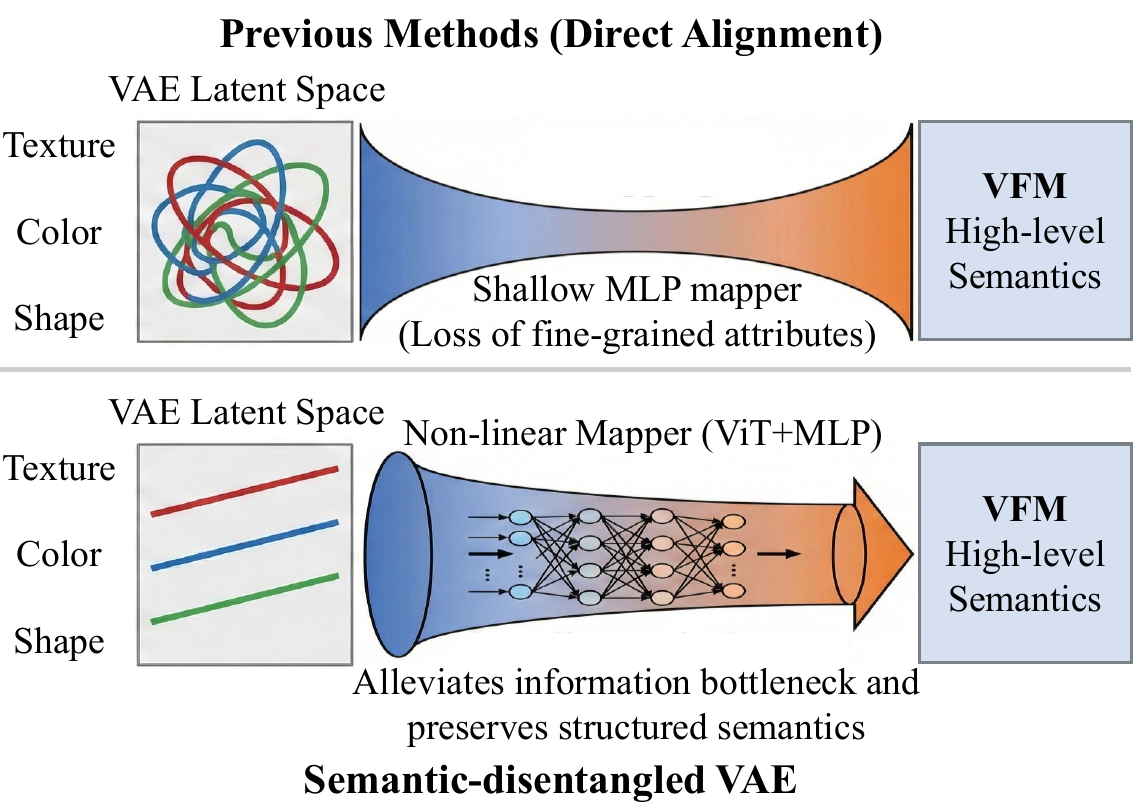}
\end{minipage}
\caption{
(Left) We reveal a strong correlation (Pearson corr. -0.9572) between the linear separability of low-level attributes (measured by Attribute Prediction Recall) in VAE latent spaces and the downstream generation quality (measured by FID). (Right) Comparison of alignment paradigms. Send-VAE introduces a non-linear mapper (ViT+MLP), which seamlessly absorbing dense VFM semantics while effectively preserving the VAE's native structured and disentangled semantics.
}
\label{fig:speed}
\vspace{-20pt}
\end{figure}

\section{Introduction}
\label{sec:intro}


Latent diffusion models (LDMs)~\cite{albergo2023building, rombach2022high, peebles2023scalable, ma2024sit} have recently achieved remarkable success in high-resolution image synthesis, establishing new benchmarks in visual fidelity and detail. A critical component of these models is the image tokenizer, typically implemented via a variational autoencoder (VAE)~\cite{kingma2013auto}. By compressing images into a structured latent space, VAEs significantly reduce the computational demands associated with high-resolution generation. Consequently, the quality of a VAE directly dictates both the training efficiency and the generative fidelity of downstream diffusion models. Despite its foundational importance, the defining characteristics of a generation-friendly VAE remain underexplored.


To bridge the gap between traditional pixel-level reconstruction and generative objectives, recent pioneering efforts \cite{yao2025reconstruction, chen2025masked, zha2025language} attempt to explicitly align VAE latents with the representations of large-scale, pre-trained Vision Foundation Models (VFMs) such as CLIP \cite{radford2021learning} or DINOv2 \cite{oquab2024dinov}. Motivated by the success of representation alignment in LDM training (\eg, REPA \cite{yu2025representation}), these approaches directly adopt similar alignment paradigms for VAEs. However, while yielding notable empirical gains, this direct inheritance harbors a critical conceptual flaw: it erroneously assumes that VAEs and LDMs share identical alignment targets, overlooking their fundamentally distinct representational demands. LDMs undoubtedly benefit from highly abstract, high-level semantics crucial for generative modeling. However, VAEs serve as the fundamental tokenizers, they must encode fine-grained, low-level visual elements (\eg, textures, colors, and local structures) in a structured way. We argue that forcing direct alignment with VFMs severely couples these low-level details, leading to feature entanglement that ultimately bottlenecks generation quality.


Drawing inspiration from recent tokenizer analyses \cite{beyer2025highly}, we hypothesize that the true catalyst for a generation-friendly VAE lies in its semantic disentanglement capabilities, which naturally enable the model to robustly encode attribute-level visual information. To validate this hypothesis, we conduct linear probing experiments on attribute prediction benchmarks to empirically quantify the disentanglement of various VAE latent spaces. As depicted in \cref{fig:speed} (left), our analysis reveals a striking positive correlation between the linear separability of low-level attributes and the final generation quality of the downstream diffusion model. This compelling evidence dictates that the richness and accessibility of structured, fine-grained semantics is a more fundamental prerequisite for VAE latents than high-level semantic alignment. Consequently, we advocate for linear probing on attribute prediction tasks as a novel, intrinsic metric for evaluating VAE latent quality.


Motivated by these insights, we introduce the Semantic-disentangled VAE (Send-VAE) to explicitly cultivate this property. Rather than enforcing the rigid, shallow alignment employed by prior works, Send-VAE leverages the rich representations of VFMs through a carefully designed non-linear mapping architecture. As illustrated in \cref{fig:speed} (right), this mapper effectively bridges the representational gap between the VAE's local structures and the VFM's dense semantics. This design facilitates the seamless injection of contextual knowledge while actively preserving the VAE's native structured semantics, thereby encouraging emergent disentanglement without requiring explicit regularization constraints.
When integrated with flow-based transformers SiTs~\cite{ma2024sit}, Send-VAE can significantly accelerate the SiT training compared with REPA and achieves a new state-of-the-art FID score of 1.21 and 1.75 with and without classifier-free guidance on ImageNet 256 $\times$ 256 generation.



In summary, this paper makes the following key contributions:

\begin{itemize}
    \item We identify semantic disentanglement as a core property of generation-friendly VAEs, verified by the strong correlation between low-level attribute prediction accuracy and downstream generative performance.
    \item We propose Send-VAE, a simple yet effective method for enhancing semantic disentanglement via alignment with vision foundation models through a sophisticated non-linear mapper network.
    \item We demonstrate that Send-VAE substantially accelerates diffusion model training and establishes new state-of-the-art results on ImageNet 256×256 generation.
\end{itemize}

\section{Related work}

\textbf{Tokenizers for Image Generation. }
Image tokenizers serve as the crucial bridge between high-dimensional pixel space and the compact latent space required by downstream generative models. These tokenizers are broadly categorized into continuous and discrete types. Continuous tokenizers, exemplified by Variational Autoencoders (VAEs)~\cite{kingma2013auto}, are widely adopted as the foundational representation in diffusion-based models~\cite{rombach2022high, peebles2023scalable, ma2024sit}. Conversely, discrete tokenizers, represented by VQGAN~\cite{esser2021taming}, dominate autoregressive (AR) generation paradigms. However, because traditional tokenizers are typically optimized with pure pixel-level reconstruction objectives, their latent spaces inevitably suffer from a semantic gap, making them not well aligned with the requirements of generation tasks.
To tackle this semantic gap, recent studies draw inspiration from advances in diffusion transformer training~\cite{yu2025representation} and attempt to explicitly align the VAE latent representations with the dense feature spaces of pre-trained VFMs. For instance, VA-VAE~\cite{yao2025reconstruction} enforces a direct alignment between VAE latents and VFM semantics, achieving significant generative performance gains while preserving reconstruction capabilities. Similarly, inspired by MAE~\cite{he2022masked}, MAETok~\cite{chen2025masked} incorporates masked image modeling into tokenizer training, leveraging multiple target features to construct a semantically rich latent space. Parallel alignment strategies have also emerged in discrete tokenizers~\cite{xiong2025gigatok, li2025imagefolder}.
While these explicit alignment strategies deliver appreciable empirical gains, they inherently adopt the identical semantic targets to both VAEs and LDMs. As we established, this direct, shallow semantic forcing severely neglects the distinct representational role of VAEs, inevitably leading to feature entanglement and the suppression of fine-grained, low-level attributes. In contrast to explicit alignment, REPA-E~\cite{leng2025repa} adopts an end-to-end joint training framework, directly backpropagating the representation alignment loss from diffusion transformers to the VAE. Although REPA-E avoids explicit feature mapping and achieves notable improvements, its black-box joint training paradigm sidesteps a more fundamental question: what intrinsic characteristics actually constitute a generation-friendly VAE?
Diverging from both rigid explicit alignment and opaque joint training, we argue that the representational needs of LDMs and VAEs differ fundamentally, and the true catalyst for a generation-friendly VAE is its semantic disentanglement capability. Rather than naively forcing VFM semantics onto VAE latents, our Send-VAE introduces a non-linear mapping mechanism. This design effectively absorbs the rich contextual guidance from VFMs while actively safeguarding the structured, fine-grained details of the VAE, thereby ensuring high-quality generation.

\noindent\textbf{Diffusion models for image generation.}
Diffusion models have emerged as a dominant paradigm in generative modeling, formulating high-fidelity image synthesis as a progressive denoising process. Early approaches, such as DDPM~\cite{ho2020denoising} and DDIM~\cite{song2021denoising}, operate directly in the high-dimensional pixel space, which heavily burdens computational resources and restricts scalability. To alleviate this, Latent Diffusion Models (LDMs)~\cite{rombach2022high} map images into a lower-dimensional, compressed latent space via pre-trained VAEs, enabling dramatically faster training and inference. Alongside this spatial compression, the field has witnessed a significant architectural paradigm shift: transitioning from the traditional U-Net backbones~\cite{nichol2021improved, rombach2022high} to transformer-based designs (e.g., DiTs)~\cite{peebles2023scalable, ma2024sit}, which are inherently more adept at capturing complex, long-range dependencies.
Beyond architectural upgrades, recent research heavily emphasizes enhancing the representation learning capabilities of the denoising network itself. Inspired by self-supervised learning, methods like MaskDiT~\cite{zheng2024fast} and SD-DiT~\cite{zhu2024sd} adapt masked modeling paradigms~\cite{he2022masked, zhouimage} to accelerate feature learning within the DiT framework. More notably, a burgeoning line of work explicitly aligns the diffusion model's intermediate features with the dense semantics of pre-trained Vision Foundation Models (VFMs). For instance, REPA~\cite{yu2025representation} leverages external semantic priors from a frozen, high-capacity encoder to regularize the generative process. Building upon this, SARA~\cite{chen2025sara} introduces additional structural and adversarial alignment objectives, while SoftREPA~\cite{lee2025aligning} extends this semantic alignment to multimodal text embeddings. Alternatively, Dispersive Loss~\cite{wang2025diffuse} demonstrates that even internal representation regularization, without relying on external VFMs, can significantly boost generative modeling.
However, these works treat the underlying VAE latent space as a static, flawless canvas and overlook the representation bottleneck within the VAE itself. Deviating from these denoising-centric alignments, our work shifts the focus back to the VAEs, proving that a semantically disentangled VAE is a crucial prerequisite for unleashing the full potential of modern diffusion models.

\section{Method}


In this section, we provide a comprehensive introduction to the design of Send-VAE. We begin by rethinking existing VAE evaluation metrics and empirically establishing that semantic disentanglement is the fundamental prerequisite for a generation-friendly VAE. To explicitly cultivate this property, we introduce Send-VAE, which incorporates a non-linear mapping architecture to safely distill semantic knowledge from VFMs without compromising the VAE's inherent structural integrity.

\begin{figure}[]
\vspace{-20pt}
\centering
\includegraphics[width=1\linewidth]{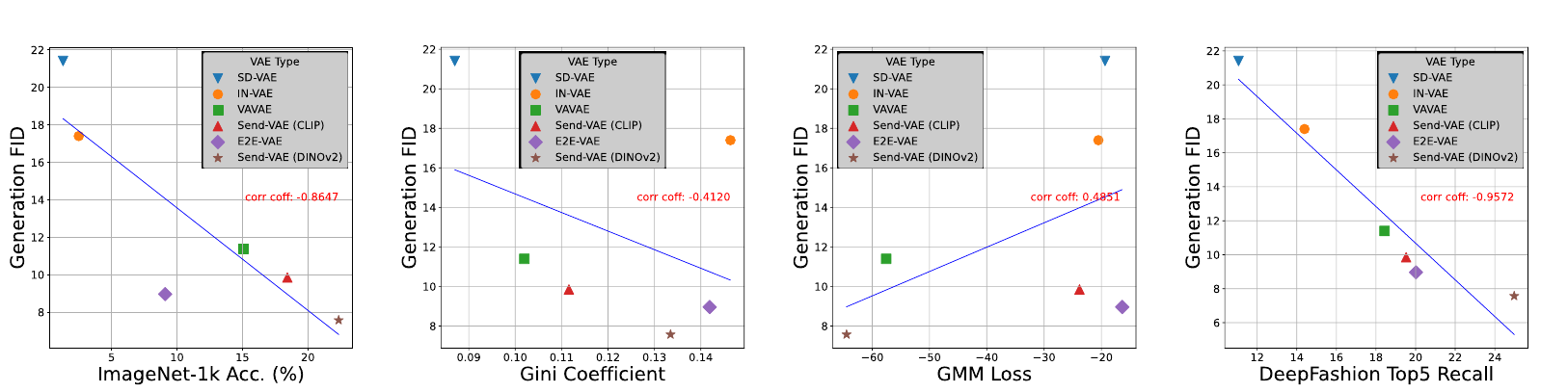}
\setlength{\abovecaptionskip}{0pt} 
\setlength{\belowcaptionskip}{-20pt}
\caption{We conduct experiments with three recently proposed evaluation methods for VAE latent space, and show their correlation with down stream generation performance (gFID). Experimental results on six VAEs with identical specifications indicate that these metrics do not accurately reflect the impact of VAEs on downstream generative performance. Conversely, we find that the ability of VAEs regarding low-level attributes is the key factor.}
\label{fig:scale}
\end{figure}

\subsection{Rethinking VAE Evaluation: The Disentanglement Hypothesis}

To answer the question of what intrinsic characteristics a generation-friendly VAE should possess, we first investigate the behavior of VAE latent spaces using three recently proposed evaluation methods: semantic gap~\cite{yu2025representation}, latent space uniformity~\cite{yao2025reconstruction}, and latent space discrimination~\cite{chen2025masked}. For the semantic gap, linear probing on ImageNet classification is adopted following REPA~\cite{yu2025representation}. Next, for latent space uniformity, we calculate the Gini coefficients of the data point distribution using kernel density estimation (KDE), as done in VA-VAE~\cite{yao2025reconstruction}. As for latent space discrimination, we fit a Gaussian mixture model (GMM) into the latent space following MAETok~\cite{chen2025masked}. We benchmark four publicly available VAEs: SD-VAE~\cite{rombach2022high}, VA-VAE (f16d32)~\cite{yao2025reconstruction}, E2E-VAE~\cite{leng2025repa}, and IN-VAE~\cite{leng2025repa}, alongside two different types of Send-VAE, with the final results presented in~\cref{fig:scale}.


\textbf{Uniformity and discrimination are insufficient indicators.} As shown in~\cref{fig:scale}, while VA-VAE exhibits improved uniformity and enhanced downstream generation performance compared with IN-VAE, this positive correlation does not hold true for E2E-VAE. A similarly inconsistent pattern emerges in the evaluation of latent space discrimination. We argue that these metrics only partially reflect the impact of VAEs on generative modeling and fail to accurately characterize a truly generation-friendly VAE.


\textbf{Semantic disentanglement is the key catalyst.} Aligning the hidden states of a diffusion model with pre-trained VFMs was first proposed in REPA~\cite{yu2025representation} to reduce the semantic gap, which accelerates convergence. However, for VAEs, we observe that while directly injecting semantic information yields partial generation improvements (\eg, VA-VAE over IN-VAE), it is not the ultimate requirement, considering the further gains achieved by E2E-VAE without explicit semantic forcing. Drawing inspiration from recent tokenizer analyses~\cite{beyer2025highly}, we hypothesize that the \textit{semantic disentanglement ability} of a VAE is the underlying key factor. To verify this, we conduct linear probing on attribute prediction tasks. As shown in~\cref{fig:scale} (right), the Top5 Recall on DeepFashion~\cite{liuLQWTcvpr16DeepFashion} dataset shows a striking positive correlation with the generation performance, strongly validating our hypothesis. Notably, our Send-VAE achieves superior semantic disentanglement, directly translating to state-of-the-art generation performance.


\subsection{The Semantic Entanglement Dilemma in Direct Alignment}
While recent explicit alignment strategies attempt to inject VFM semantics into VAEs, they predominantly rely on shallow mapping networks, such as a simple multilayer perceptron (MLP) used in VA-VAE~\cite{yao2025reconstruction}. However, this direct alignment paradigm harbors a critical conceptual flaw. Pre-trained VFMs (\eg, DINOv2) inherently encode highly condensed, globalized, and deeply entangled high-level semantics. Conversely, a foundational VAE must preserve fine-grained, disentangled local visual elements (\eg, textures, colors, and structures) to ensure high-fidelity tokenization. Forcing a rigid, shallow alignment between these two fundamentally mismatched representational spaces inevitably compels the VAE to discard its delicate structural details to accommodate the VFM's dense representations. This indiscriminate semantic forcing overwhelmingly couples low-level features, leading to severe semantic collapse and feature entanglement, which ultimately bottlenecks the downstream generative modeling.

\subsection{Semantic Disentangled VAE}
To overcome the aforementioned dilemma and actively enhance the disentanglement capability, we propose the Semantic-disentangled VAE (Send-VAE). Deviating from the direct alignment paradigm, Send-VAE introduces a sophisticated non-linear mapper network to bridge the substantial representational gap between the VAE and VFMs.

\begin{wrapfigure}{r}{0.5\textwidth} 
\vspace{-20pt}
\setlength{\abovecaptionskip}{0pt} 
\setlength{\belowcaptionskip}{-10pt}
  \centering
  \includegraphics[width=0.5\textwidth]{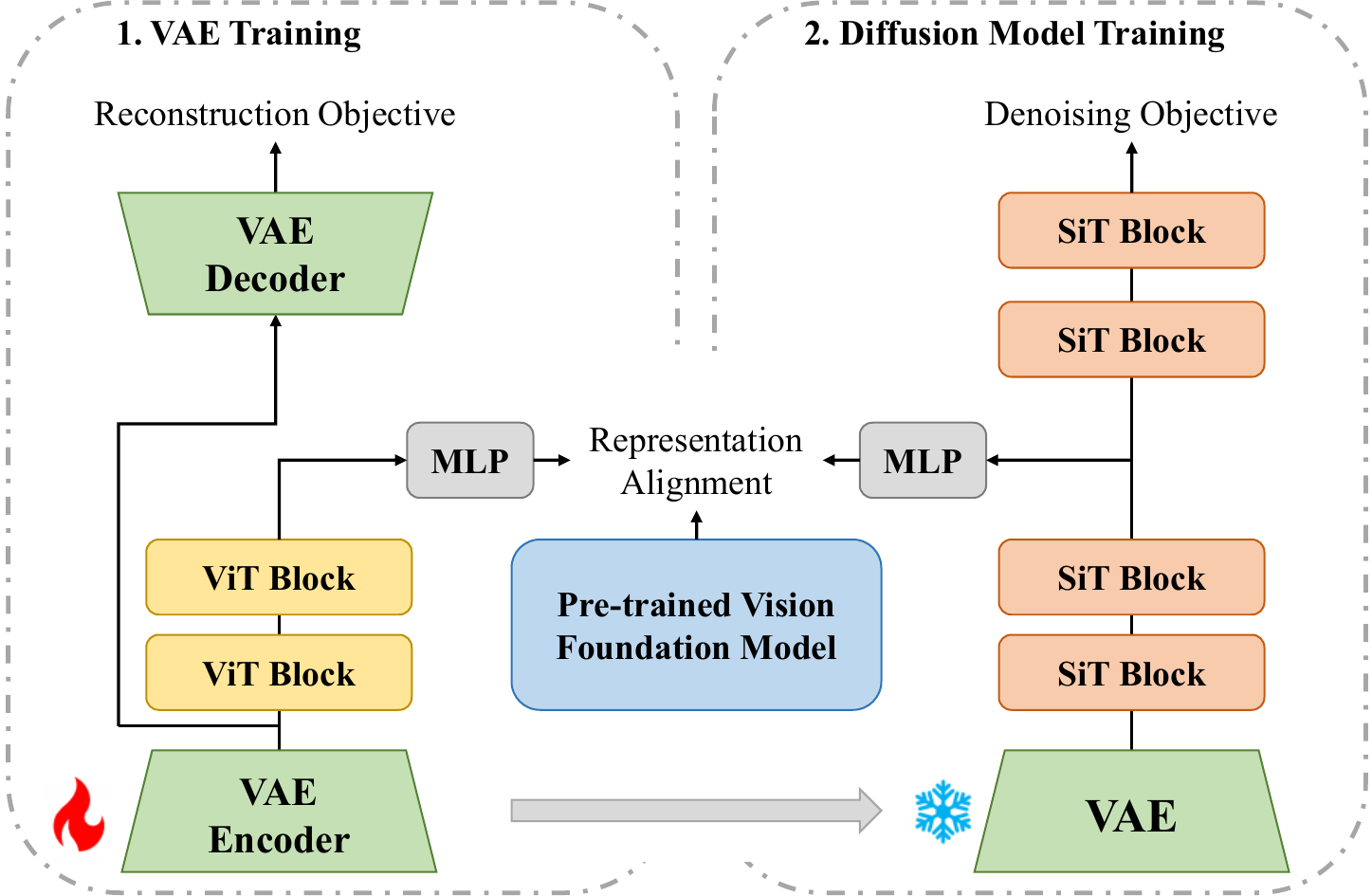}
  \caption{The overall structure of Send-VAE.}
  \label{fig:side_image}
\end{wrapfigure}

Specifically, rather than utilizing a simple MLP mapper, our mapper network consists of a patch embedding layer, a stack of Vision Transformer (ViT)~\cite{dosovitskiy2021an} layers, and a final MLP projector. This deep architecture acts as an essential \textit{semantic buffer}. It possesses sufficient capacity to digest and translate the dense, entangled semantics from the VFM into a structured format that the VAE can safely absorb. Consequently, this non-linear buffering facilitates the seamless injection of valuable contextual knowledge while actively safeguarding the VAE's native structured semantics, naturally encouraging emergent disentanglement without explicit rigid regularization. The overall framework is illustrated in~\cref{fig:side_image}.

Formally, given a clean image $\mathbf{x}$, let $\mathbf{z}$ be the latent representation of $\mathbf{x}$ output by the VAE $\mathcal{V}_{\theta}$. Let $f$ denote a frozen Vision Foundation Model, where $\mathbf{y}=f(\mathbf{x})\in\mathbb{R}^{N\times D}$ is the encoded representation of $\mathbf{x}$, with $N$ and $D$ being the number of patches and the embedding dimension of $f$, respectively. To ensure the learned latent space is strictly robust and aligned with the actual requirements of downstream generative modeling, we propose a diffusion-aware alignment strategy. Instead of aligning clean latent codes, Send-VAE simulates the input distribution of downstream diffusion models by injecting random Gaussian noise into $\mathbf{z}$ to obtain the noisy latent $\mathbf{z}_t$:

\begin{equation}
    \begin{aligned}
        \mathbf{z}_t = (1 - \alpha_t)\epsilon + \alpha_t\mathbf{z}; \epsilon \in \mathcal{N}(\mathbf{0}, \mathbf{I}),
    \end{aligned}
\end{equation}
where $\alpha_t$ is randomly sampled from a uniform distribution between 0 and 1 and $t$ is the time step (following the formulation in SiT~\cite{ma2024sit}).
Subsequently, the non-linear mapper network $h_{\phi}$ transforms $\mathbf{z}_t$ into $h_\phi(\mathbf{z}_t)$. The alignment loss is then calculated using the patch-wise cosine similarity between the transformed latent and the VFM features:
\begin{equation}
    \begin{aligned}
        \mathcal{L}_{\text{align}}=\frac{1}{N}\sum_{n=1}^N \left(1-\frac{h_\phi(\mathbf{z}_t)^{[n]}\cdot f(\mathbf{x})^{[n]}}{\lVert h_\phi(\mathbf{z}_t)^{[n]}\rVert\lVert f(\mathbf{x})^{[n]}\rVert}\right),
    \end{aligned}
\end{equation}
where $n$ is the patch index.

In practice, we utilize $\mathcal{L}_{\text{align}}$ to fine-tune a pre-trained VAE for fast convergence. The original VAE training objective $\mathcal{L}_{\text{VAE}}$~\cite{stabilityai2025sdvae} is simultaneously maintained to preserve foundational reconstruction capabilities, which comprises reconstruction losses ($\mathcal{L}_{\text{MSE}}$, $\mathcal{L}_{\text{LPIPS}}$), adversarial GAN loss ($\mathcal{L}_{\text{GAN}}$), and KL divergence loss ($\mathcal{L}_{\text{KL}}$). Thus, the overall training objective is formulated as:
\begin{equation}
    \begin{aligned}
        \mathcal{L}(\theta, \phi)=\lambda_{\text{align}}\mathcal{L}_{\text{align}} + \mathcal{L}_{\text{VAE}},
    \end{aligned}
\end{equation}
where $\theta$ and $\phi$ refer to the trainable parameters of the VAE and the non-linear mapper network, respectively.

\section{Experiments}

In this section, we conduct comprehensive experiments on the ImageNet dataset~\cite{deng2009imagenet} at 256×256 resolution to validate the design choices of Send-VAE, and benchmark its generation performance to demonstrate its superiority over existing approaches.

\subsection{Implementation Details}

We follow the same set up as in REPA-E~\cite{leng2025repa} unless otherwise specified.
All training is conducted on the training split of ImageNet~\cite{deng2009imagenet}.
The data preprocessing protocol is same as in ADM~\cite{dhariwal2021diffusion} including center-crop and resizing to 256x256 resolution. The mapper network consists of one layer of Transformer encoder with 12 heads for VFMs with 768 hidden dimension and 16 heads for VFMs with 1024 hidden dimension.

\textbf{For VAE training}, we train 80 epoch with a global batch size of 1024, AdamW~\cite{loshchilov2018decoupled} optimizer is adopted and the learning rate is set to $3.0\times10^{-4}$.
As for the initialization, we experiment with publicly available VAEs, including SD-VAE (f8d4)~\cite{rombach2022high}, VA-VAE (f16d32)~\cite{yao2025reconstruction}, and IN-VAE (f16d32), which is trained on ImageNet following~\cite{rombach2022high}.
Experimentally, we choose VA-VAE as the default setting.
As for alignment loss $\mathcal{L}_{\text{align}}$, we use DINOv2~\cite{oquab2024dinov} as the vision foundation model, and $\lambda_{\text{align}}$ is set to 1.0.

\textbf{For diffusion models}, we choose SiT-XL/1 and SiT-XL/2 for VAEs with 4× and 16× downsampling rates, respectively, where
1 and 2 denote the patch sizes in the transformer embedding layer.
We train either 80 epoch or 800 epoch with a global batch size of 256, and gradient clipping and exponential moving average
(EMA) are applied stable optimization.
The learning rate is set to $1.0\times10^{-4}$ and AdamW optimizer is used.
REPA loss is also included following the setting in~\cite{yu2025representation}.

\textbf{For sampling}, the SDE Euler-Maruyama sampler is used, the number of function evaluations (NFE) is set to 250 by default and the cfg scale is set to 2.5

\subsection{Evaluation Metrics}

For image generation evaluation, we strictly follow the ADM setup~\cite{dhariwal2021diffusion}. Generation quality is assessed using Fréchet Inception Distance (gFID)~\cite{heusel2017gans}, Structural FID (sFID)~\cite{nash2021generating}, Inception Score (IS)~\cite{salimans2016improved}, Precision, and Recall~\cite{kynkaanniemi2019improved}, computed on 50K generated samples. For sampling, we adopt the SDE Euler–Maruyama solver with 250 steps, following the protocols of REPA~\cite{yu2025representation} and REPA-E~\cite{leng2025repa}. For VAE evaluation, we report reconstruction FID (rFID) on 50K validation images from ImageNet at 256×256 resolution.

\begin{table}[t]
\centering
\setlength{\belowcaptionskip}{0pt}
\setlength{\tabcolsep}{1pt}
\renewcommand\arraystretch{1.5}
\caption{System-level comparison on ImageNet 256x256 generation with and without classifier-free guidance (CFG). Our Send-VAE can significant accelerate the convergence of diffusion models, which achieves a gFID socre of 2.88/1.41 wo/w CFG for only 80 epoch of training. Although the performance gap between Send-VAE and E2E-VAE is narrowing when training longer, Send-VAE still achieves further improvements.}
\label{tab:sota}
\resizebox{1\textwidth}{!}{
\begin{tabular}{ccccccccccccccc}
\hline
\multicolumn{1}{c|}{\multirow{2}{*}{Tokenizer}}       & \multicolumn{1}{c|}{\multirow{2}{*}{Method}}      & \multicolumn{1}{c|}{\multirow{2}{*}{\begin{tabular}[c]{@{}c@{}}Training \\ Epoch\end{tabular}}} & \multicolumn{1}{c|}{\multirow{2}{*}{\#params}} & \multicolumn{1}{c|}{\multirow{2}{*}{rFID}} & \multicolumn{5}{c|}{Generation w/o CFG}                  & \multicolumn{5}{c}{Generation w/ CFG} \\ \cline{6-15} 
\multicolumn{1}{c|}{}                                 & \multicolumn{1}{c|}{}                             & \multicolumn{1}{c|}{}                                                                           & \multicolumn{1}{c|}{}                          & \multicolumn{1}{c|}{}                      & gFID & sFID  & IS    & Prec. & \multicolumn{1}{c|}{Rec.} & gFID  & sFID  & IS     & Prec. & Rec. \\ \hline
\multicolumn{15}{c}{AutoRegressive (AR)}                                                                                                                                                                                                                                                                                                                                                                     \\ \hline
\multicolumn{1}{c|}{MaskGiT}                          & \multicolumn{1}{c|}{MaskGIT~\cite{chang2022maskgit}}                      & \multicolumn{1}{c|}{555}                                                                        & \multicolumn{1}{c|}{227M}                      & \multicolumn{1}{c|}{2.28}                  & 6.18 & -     & 182.1 & 0.80  & \multicolumn{1}{c|}{0.51} & -     & -     & -      & -     & -    \\
\multicolumn{1}{c|}{VQGAN}                            & \multicolumn{1}{c|}{LlamaGen~\cite{sun2024autoregressive}}                     & \multicolumn{1}{c|}{300}                                                                        & \multicolumn{1}{c|}{3.1B}                      & \multicolumn{1}{c|}{0.59}                  & 9.38 & 8.24  & 112.9 & 0.69  & \multicolumn{1}{c|}{0.67} & 2.18  & 5.97  & 263.3  & 0.81  & 0.58 \\
\multicolumn{1}{c|}{VQVAE}                            & \multicolumn{1}{c|}{VAR~\cite{tian2024visual}}                          & \multicolumn{1}{c|}{350}                                                                        & \multicolumn{1}{c|}{2.0B}                      & \multicolumn{1}{c|}{-}                     & -    & -     & -     & -     & \multicolumn{1}{c|}{-}    & 1.80  & -     & 365.4  & 0.83  & 0.57 \\
\multicolumn{1}{c|}{LFQ tokenizers}                   & \multicolumn{1}{c|}{MagViT-v2~\cite{yu2024language}}                    & \multicolumn{1}{c|}{1080}                                                                       & \multicolumn{1}{c|}{307M}                      & \multicolumn{1}{c|}{1.50}                  & 3.65 & -     & 200.5 & -     & \multicolumn{1}{c|}{-}    & 1.78  & -     & 319.4  & -     & -    \\
\multicolumn{1}{c|}{LDM}                              & \multicolumn{1}{c|}{MAR~\cite{li2024autoregressive}}                          & \multicolumn{1}{c|}{800}                                                                        & \multicolumn{1}{c|}{945M}                      & \multicolumn{1}{c|}{0.53}                  & 2.35 & -     & 227.8 & 0.79  & \multicolumn{1}{c|}{0.62} & 1.55  & -     & 303.7  & 0.81  & 0.62 \\ \hline
\multicolumn{15}{c}{Latent Diffusion Models (LDM)}                                                                                                                                                                                                                                                                                                                                                           \\ \hline
\multicolumn{1}{c|}{\multirow{7}{*}{SD-VAE~\cite{rombach2022high}}}          & \multicolumn{1}{c|}{MaskDiT~\cite{zheng2024fast}}                      & \multicolumn{1}{c|}{1600}                                                                       & \multicolumn{1}{c|}{675M}                      & \multicolumn{1}{c|}{\multirow{7}{*}{0.61}} & 5.69 & 10.34 & 177.9 & 0.74  & \multicolumn{1}{c|}{0.60} & 2.28  & 5.67  & 276.6  & 0.80  & 0.61 \\
\multicolumn{1}{c|}{}                                 & \multicolumn{1}{c|}{DiT~\cite{peebles2023scalable}}                          & \multicolumn{1}{c|}{1400}                                                                       & \multicolumn{1}{c|}{675M}                      & \multicolumn{1}{c|}{}                      & 9.62 & 6.85  & 121.5 & 0.67  & \multicolumn{1}{c|}{0.67} & 2.27  & 4.60  & 278.2  & 0.83  & 0.57 \\
\multicolumn{1}{c|}{}                                 & \multicolumn{1}{c|}{SiT~\cite{ma2024sit}}                          & \multicolumn{1}{c|}{1400}                                                                       & \multicolumn{1}{c|}{675M}                      & \multicolumn{1}{c|}{}                      & 8.61 & 6.32  & 131.7 & 0.68  & \multicolumn{1}{c|}{0.67} & 2.06  & 4.50  & 270.3  & 0.82  & 0.59 \\
\multicolumn{1}{c|}{}                                 & \multicolumn{1}{c|}{FastDiT~\cite{yao2024fasterdit}}                      & \multicolumn{1}{c|}{400}                                                                        & \multicolumn{1}{c|}{675M}                      & \multicolumn{1}{c|}{}                      & 7.91 & 5.45  & 131.3 & 0.67  & \multicolumn{1}{c|}{0.69} & 2.03  & 4.63  & 264.0  & 0.81  & 0.60 \\
\multicolumn{1}{c|}{}                                 & \multicolumn{1}{c|}{MDT~\cite{gao2023masked}}                          & \multicolumn{1}{c|}{1300}                                                                       & \multicolumn{1}{c|}{675M}                      & \multicolumn{1}{c|}{}                      & 6.23 & 5.23  & 143.0 & 0.71  & \multicolumn{1}{c|}{0.65} & 1.79  & 4.57  & 283.0  & 0.81  & 0.61 \\
\multicolumn{1}{c|}{}                                 & \multicolumn{1}{c|}{MDTv2~\cite{gao2023mdtv2}}                        & \multicolumn{1}{c|}{1080}                                                                       & \multicolumn{1}{c|}{675M}                      & \multicolumn{1}{c|}{}                      & -    & -     & -     & -     & \multicolumn{1}{c|}{-}    & 1.58  & 4.52  & 314.7  & 0.79  & 0.65 \\
\multicolumn{1}{c|}{}                                 & \multicolumn{1}{c|}{REPA~\cite{yu2025representation}}                         & \multicolumn{1}{c|}{800}                                                                        & \multicolumn{1}{c|}{675M}                      & \multicolumn{1}{c|}{}                      & 5.90 & 5.73  & 157.8 & 0.70  & \multicolumn{1}{c|}{0.69} & 1.42  & 4.70  & 305.7  & 0.80  & 0.65 \\ \hline
\multicolumn{1}{c|}{\multirow{2}{*}{VA-VAE~\cite{yao2025reconstruction}}}          & \multicolumn{1}{c|}{\multirow{2}{*}{LightingDiT~\cite{yao2025reconstruction}}} & \multicolumn{1}{c|}{80}                                                                         & \multicolumn{1}{c|}{675M}                      & \multicolumn{1}{c|}{0.28}                  & 4.29 & -     & -     & -     & \multicolumn{1}{c|}{-}    & -     & -     & -      & -     & -    \\
\multicolumn{1}{c|}{}                                 & \multicolumn{1}{c|}{}                             & \multicolumn{1}{c|}{800}                                                                        & \multicolumn{1}{c|}{675M}                      & \multicolumn{1}{c|}{0.28}                  & 2.17 & 4.36  & 205.6 & 0.77  & \multicolumn{1}{c|}{0.65} & 1.35  & 4.15  & 295.3  & 0.79  & 0.65 \\ \hline
\multicolumn{1}{c|}{MAETok~\cite{chen2025masked}}                           & \multicolumn{1}{c|}{LightingDiT~\cite{yao2025reconstruction}}                  & \multicolumn{1}{c|}{800}                                                                        & \multicolumn{1}{c|}{675M}                      & \multicolumn{1}{c|}{0.48}                  & 2.21 & -     & 208.3 & -     & \multicolumn{1}{c|}{-}    & 1.73  & -     & 308.4  &   -   &   -  \\ \hline
\multicolumn{1}{c|}{\multirow{2}{*}{E2E-VAE~\cite{leng2025repa}}}         & \multicolumn{1}{c|}{\multirow{2}{*}{REPA~\cite{yu2025representation}}}        & \multicolumn{1}{c|}{80}                                                                         & \multicolumn{1}{c|}{675M}                      & \multicolumn{1}{c|}{\multirow{2}{*}{0.28}} & 3.46 & 4.17  & 159.8 & 0.77  & \multicolumn{1}{c|}{0.63} & 1.67  & 4.12  & 266.3  & 0.80  & 0.63 \\
\multicolumn{1}{c|}{}                                 & \multicolumn{1}{c|}{}                             & \multicolumn{1}{c|}{800}                                                                        & \multicolumn{1}{c|}{675M}                      & \multicolumn{1}{c|}{}                      & 1.83 & \textbf{4.22}  & 217.3 & 0.77  & \multicolumn{1}{c|}{0.66} & 1.26  & 4.11  & 314.9  & 0.79  & 0.66 \\ \hline
\multicolumn{1}{c|}{\multirow{2}{*}{\makecell{Send-VAE}}}         & \multicolumn{1}{c|}{\multirow{2}{*}{REPA~\cite{yu2025representation}}}        & \multicolumn{1}{c|}{80}                                                                         & \multicolumn{1}{c|}{675M}                      & \multicolumn{1}{c|}{\multirow{2}{*}{0.31}} & 2.88 & 4.67  & 175.3 & 0.78  & \multicolumn{1}{c|}{0.62} & 1.41  & 4.41  & 301.7  & 0.79  & 0.65 \\
\multicolumn{1}{c|}{}                                 & \multicolumn{1}{c|}{}                             & \multicolumn{1}{c|}{800}                                                                        & \multicolumn{1}{c|}{675M}                      & \multicolumn{1}{c|}{}                      & \textbf{1.75} & 4.41  & 218.57 & \textbf{0.79}  & \multicolumn{1}{c|}{0.64} & \textbf{1.21}  & \textbf{4.10}  & \textbf{315.1}  & \textbf{0.79}  & \textbf{0.66} \\ \hline
\end{tabular}}
\vspace{-20pt}
\end{table}

\subsection{System-level comparison on ImageNet 256x256 generation}

To verify the effectiveness of Send-VAE, we conduct system-level comparison on ImageNet 256x256 generation with and without classifier-free guidance (CFG), and present the results in~\cref{tab:sota}.
As we can see, using the same vision foundation model DINOV2, Send-VAE can achieve notable performance gains compared with E2E-VAE and set a new state-of-the-art generation FID score of 1.21 and 1.75 with and without classifier-free guidance on ImageNet 256x256 generation.
These results highly demonstrate the effectiveness of enhancing the semantic
\begin{wrapfigure}{r}{0.3\textwidth} 
\vspace{-20pt}
\setlength{\abovecaptionskip}{0pt} 
\setlength{\belowcaptionskip}{-20pt}
  \centering
  \includegraphics[width=0.3\textwidth]{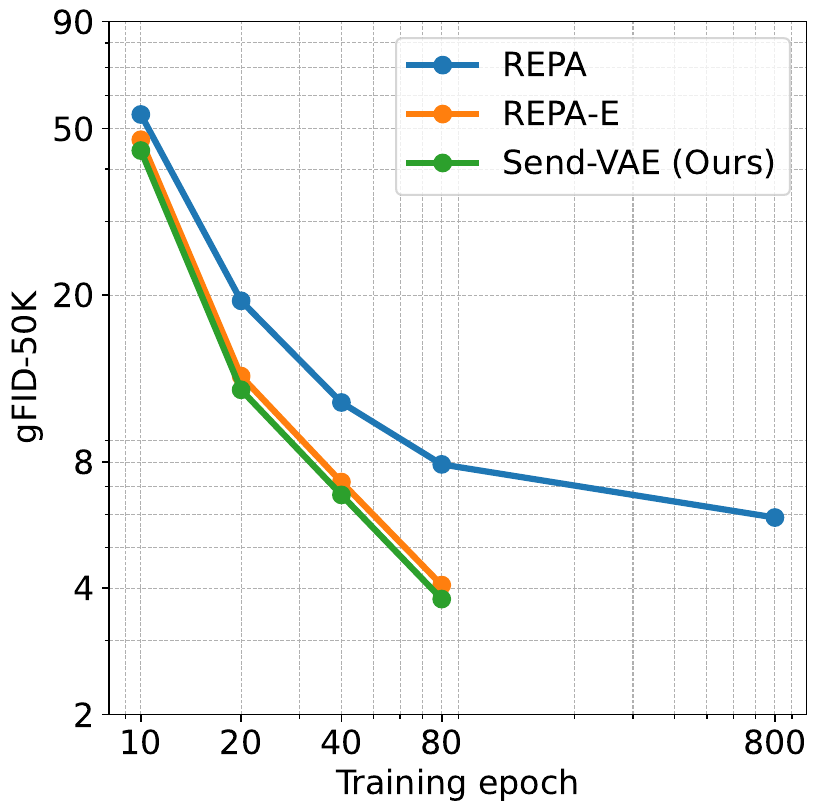}
  \caption{Send-VAE can significantly accelerate the SiT training.}
  \label{fig:speed_exp}
\end{wrapfigure}
 disentanglement ability of VAE.
Meanwhile, we can notice that Send-VAE can significantly speed up the convergence of diffusion models compared with REPA (shown in ~\cref{fig:speed_exp}).
When compared with REPA-E, Send-VAE can narrow the gFID score from 3.46 to 2.88 for generation without CFG when training with only 80 epoch, which demonstrates that Send-VAE is a generation-friendly VAE and can facilitate the learning of diffusion models.
As for reconstruction performance, we observe that the reconstruction performance of Send-VAE is slightly inferior to that of VA-VAE. We attribute this to the semantic disentangled latent space of Send-VAE, which prevents it from capturing excessive fine-grained low-level details.

\begin{figure}[t]
    \centering
    \setlength{\abovecaptionskip}{0pt} 
    \setlength{\belowcaptionskip}{0pt}
    \includegraphics[width=1.0\linewidth]{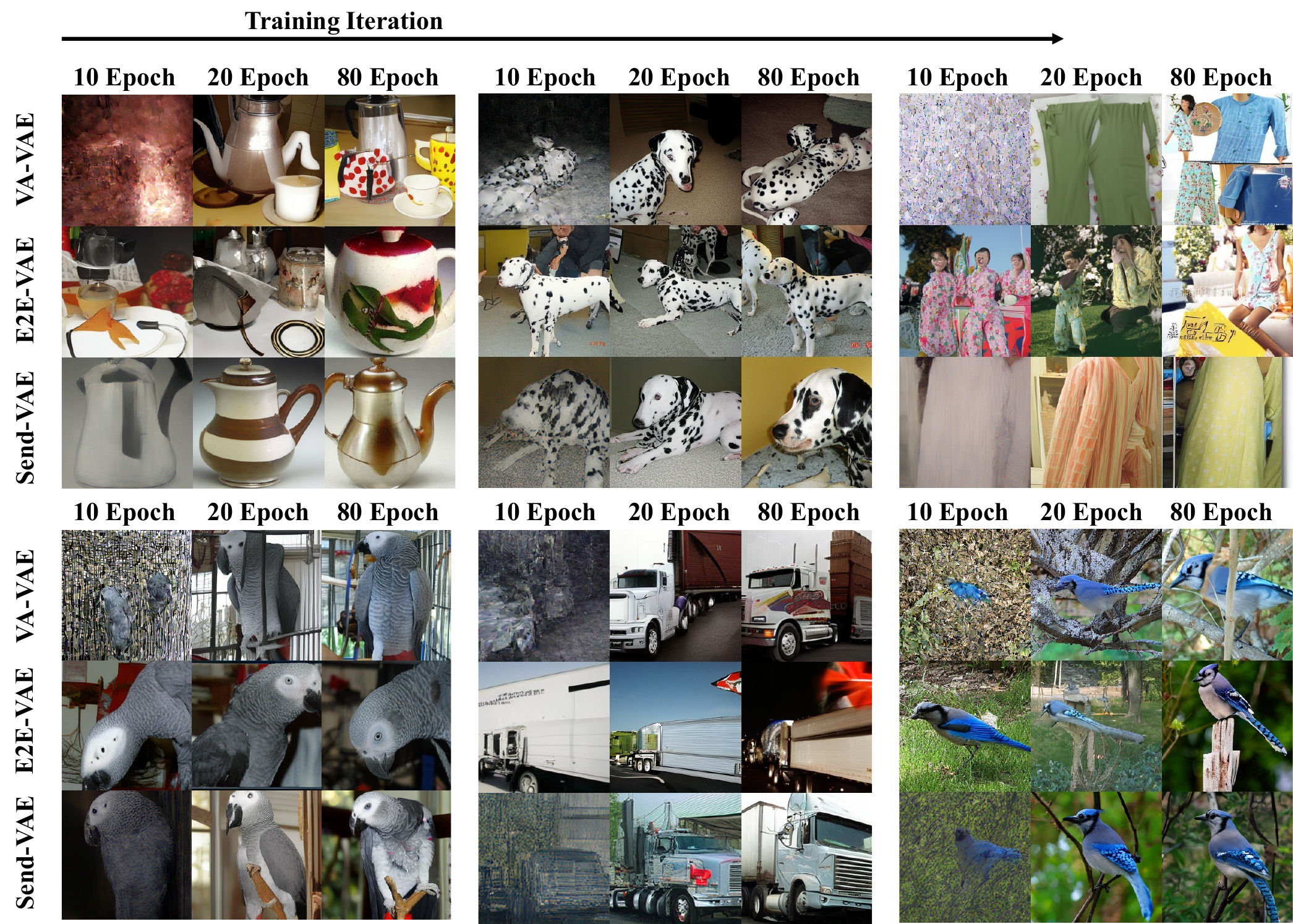}
    \caption{Qualitative comparisons among VA-VAE, E2E-VAE, and Send-VAE. Results for both methods are sampled using the same seed, noise and class label. The classifier-free guidance scale is set to 4.0.}
    \label{fig:epoch}
    \vspace{-15pt}
\end{figure}

\begin{figure}[h]
    \centering
    \setlength{\abovecaptionskip}{0pt} 
    \setlength{\belowcaptionskip}{0pt}
    \includegraphics[width=1.0\linewidth]{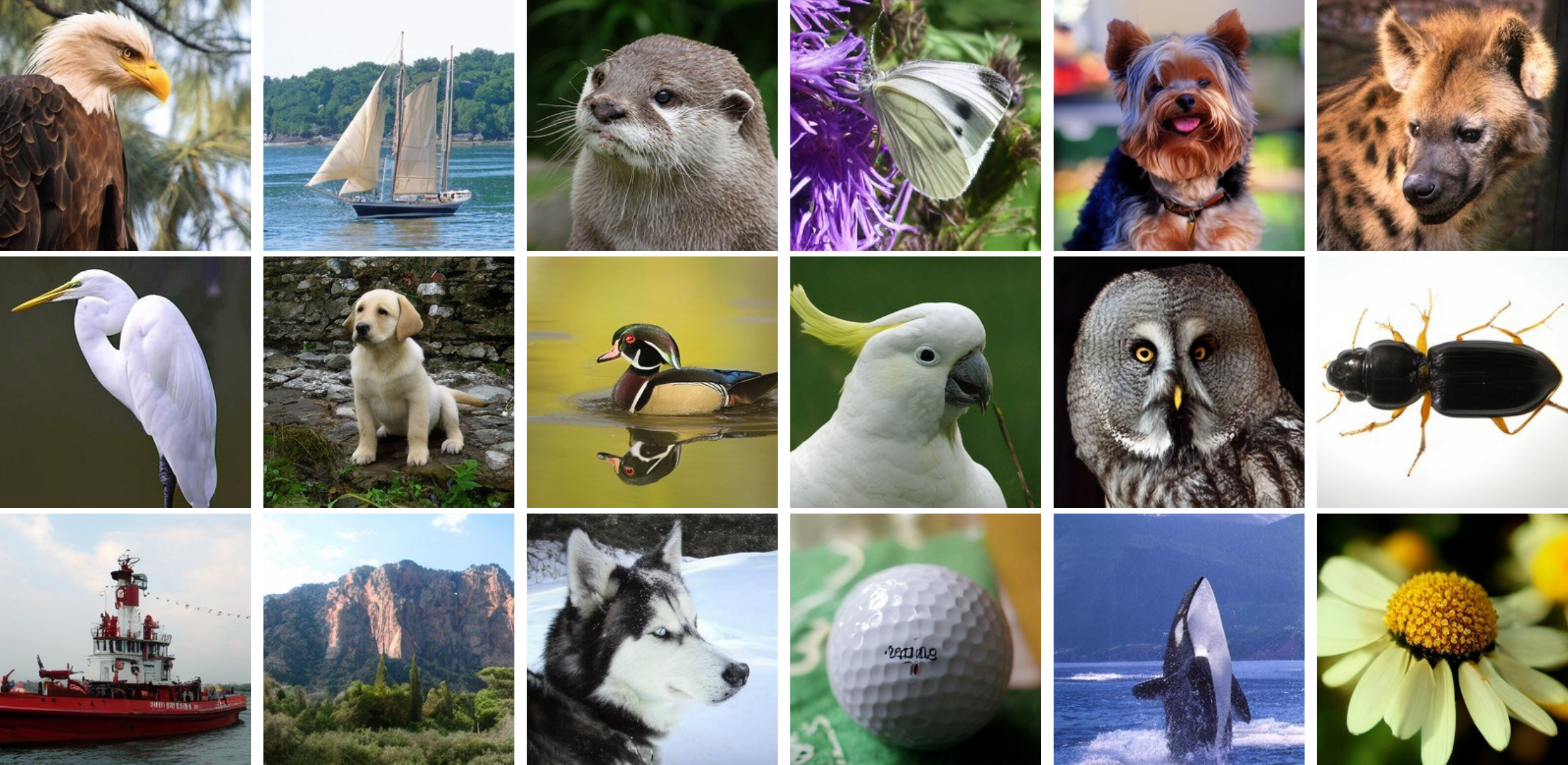}
    \caption{Qualitative Results on ImageNet 256 × 256 using Send-VAE and SiT-XL.}
    \label{fig:vis}
\end{figure}

Besides, we also provide qualitative comparisons among VA-VAE, E2E-VAE and Send-VAE in~\cref{fig:epoch} 
We generates images from the same label and initial noise using checkpoints trained by 10 epoch, 20 epoch, and 80 epoch, respectively.
As we can see, training diffusion models using Send-VAE demonstrates superior image generation quality compared to VA-VAE and E2E-VAE.
Meanwhile, Send-VAE can significantly speed up the training process of diffusion models, evidenced by the more structurally meaningful images during early stages of training process.
Some visualization results are presented in~\cref{fig:vis} using Send-VAE and SiT-XL/1 to show that training diffusion models with Send-VAE can generate high-quality images.

\subsection{Ablation Studies}

In this section, we provide detailed ablation studies to demonstrate the effectiveness of each design in Send-VAE. 
Unless otherwise specified, we train a SiT-B/1 with REPA loss for 80 epoch, and report the downstream generation performance without classifier-free guidance.

\begin{table}[]
\setlength{\abovecaptionskip}{0pt} 
\setlength{\belowcaptionskip}{0pt}
\begin{minipage}{.53\textwidth}
\centering
\small
\setlength{\tabcolsep}{3pt}
\renewcommand\arraystretch{1.5}
\caption{Ablation on the depth of mapper network.}
\resizebox{0.9\linewidth}{!}{
\begin{tabular}{cccccc}
\toprule
Depth & gFID$\downarrow$ & sFID$\downarrow$ & IS$\uparrow$ & Prec.$\uparrow$ & Rec.$\uparrow$ \\ \midrule
0              &  9.20 &  7.06 & 104.2 &  0.73  &  0.57 \\
1              &  \textbf{8.42}  &  \textbf{5.05}  & \textbf{108.3} &  \textbf{0.74}  & \textbf{0.60}    \\
2          &  9.47 &  5.33 & 100.4 & 0.73 &  \textbf{0.60}    \\ \bottomrule
\end{tabular}
}
\label{tab:ini}
\end{minipage}
\ 
\begin{minipage}{.47\textwidth}
\centering
\setlength{\tabcolsep}{3pt}
\renewcommand\arraystretch{1.5}
\caption{Ablation on noise injection.}
\resizebox{1\linewidth}{!}{
\begin{tabular}{cccccc}
\toprule
\makecell{Noise\\Injection} & gFID$\downarrow$ & sFID$\downarrow$ & IS$\uparrow$ & Prec.$\uparrow$ & Rec.$\uparrow$ \\ \midrule
\ding{55}      &    8.42  &  5.05  & 108.3 &  0.74  & 0.60 \\
\ding{51}          &   \textbf{7.57}   & 5.37 & 115.3 &   \textbf{0.74}    &   \textbf{0.60}   \\ \bottomrule
\end{tabular}
}
\label{tab:noise}
\end{minipage}
\vspace{-10pt}
\end{table}

\textbf{Ablation on Depth of Mapper Network.}
We ablate the depth of our proposed mapper network to analyze its impact on downstream generation performance. As shown in~\cref{tab:ini}, a mapper with one layer of ViT achieves the best performance (gFID=8.42), outperforming both shallower (0 layer) and deeper (2 layer) configurations.
We argue that the insufficient capacity of shallow mapper fails to bridge the representation gap between VAE and visual foundation models, resulting in a decrease in the semantic disentanglement ability of VAE.
While for the deeper one, it weaken the foundational model’s impacts on VAE due to the stronger fitting capability.
Such experimental results demonstrate the necessity of employing a mapper network to bridge representation gap, which can facilitate effective semantic injection.

\textbf{Ablation on Injecting Noise to Latent Representations.}~\cref{tab:noise} presents the ablation results of injecting noise to latent representations.
As we can see, injecting noise during the alignment process can bring significant performance gains.
We attribute its effectiveness to a form of data augmentation, which ensures that even with noise injected, the latent representation extracted by the VAE retains rich disentangled semantic information, making it better suited for the denoising process of the downstream diffusion model.

\begin{table}[t]
\vspace{-5pt}
\centering
\setlength{\abovecaptionskip}{0pt} 
\setlength{\belowcaptionskip}{0pt}
\setlength{\tabcolsep}{9pt}
\renewcommand\arraystretch{1.5}
\caption{Ablation on different vision foundation models (VFMs)}
\label{tab:vfm}
\begin{tabular}{cccccc}
\toprule
VFMs   & gFID$\downarrow$ & sFID$\downarrow$ & IS$\uparrow$ & Prec.$\uparrow$ & Rec.$\uparrow$  \\ \midrule
None (Baseline)   & 11.40 & 6.58 &  93.5  &    0.71   &  0.59    \\ 
MAE   & 10.01 & 5.62 &  99.2  &    0.71   &  0.60    \\ 
CLIP   & 9.85 & 5.59 &  100.8  &    0.71   &  0.62    \\ 
I-JEPA & 9.70 & 5.40 &  102.9  &   0.72    &  0.60    \\ 
SigLIP & 9.10 & 5.21 &  108.1  &   0.72    &  0.61s    \\ 
DINOv2 & 7.57 & 5.37 &  115.3  &  0.74 &  0.60 \\ 
DINOv3 & 7.16 & 5.57 &  125.3  & 0.75 &  0.58 \\ \bottomrule
\end{tabular}
\vspace{-10pt}
\end{table}

\textbf{Ablation on Vision Foundation Models.}
We also investigate the influence of vision foundation models and present the ablation results in~\cref{tab:vfm}.
Specifically, we include six types of vision foundation models, including MAE~\cite{he2022masked}, CLIP~\cite{radford2021learning}, I-JEPA~\cite{assran2023self}, SigLIP~\cite{zhai2023sigmoid}, DINOv2~\cite{oquab2024dinov}, and DINOv3~\cite{simeoni2025dinov3}.
As we can see, regardless of the type of vision foundation models, adding $\mathcal{L}_{\text{align}}$ consistently improve the generation performance of diffusion models.
Among them, the DINO family (DINOv2 and DINOv3) achieves the best performance, which is consistent with the findings of REPA and REPA-E.
We argue that the object-centric features of DINO can more effectively facilitate the VAE in learning a semantic disentangled latent space, thus resulting in superior generation performance.

\begin{table}[]
\vspace{-10pt}
\centering
\setlength{\tabcolsep}{6pt}
\renewcommand\arraystretch{1.5}
\setlength{\belowcaptionskip}{0pt}
\caption{Ablation on the VAE specification.}
\label{tab:vae_spec}
\begin{tabular}{ccccccc}
\toprule
VAE Specification & rFID$\downarrow$ & gFID$\downarrow$ & sFID$\downarrow$ & IS$\uparrow$ & Prec.$\uparrow$ & Rec.$\uparrow$ \\ \midrule
f16d16          &  0.48 & 13.85 &  5.30 &  65.0  &  0.68  & \textbf{0.62}  \\
\ \ \ \ \ \ +$\mathcal{L}_{\text{align}}$     & \textbf{0.47}         &   \textbf{7.02}   &  \textbf{4.64} &  \textbf{120.41}  &   \textbf{0.75}    &   0.60   \\ \midrule
f16d32          & \textbf{0.26}   & 17.43 &  5.93    &  72.7  &  0.64  &  \textbf{0.63}  \\
\ \ \ \ \ \ +$\mathcal{L}_{\text{align}}$       & 0.28       &   \textbf{8.25} &  \textbf{4.68} & \textbf{105.2} &  \textbf{0.74}  &  0.60   \\ \midrule
f16d64         & \textbf{0.17}   &  26.27 &  8.02    &  54.69  &  0.58  &  \textbf{0.64} \\
\ \ \ \ \ \ +$\mathcal{L}_{\text{align}}$    &   0.19       &  \textbf{10.99} &  \textbf{5.17} & \textbf{94.11} &  \textbf{0.69}  &  0.62    \\ \bottomrule
\end{tabular}
\vspace{-15pt}
\end{table}

\textbf{Ablation on the VAE specification.}
We implement our Send-VAE cross different VAE specification and show the results in~\cref{tab:vae_spec}. It can be observed that under varying VAE latent dimensions, Send-VAE consistently achieves significant performance improvements, demonstrating the effectiveness of our method. Furthermore, while the reconstruction quality improves with increased VAE model capacity, the generation quality reduces. And the integration of Send-VAE stably enhances the downstream generation performance.

\textbf{Ablation on the Resolution and Initialization of VAE.}
To demonstrate the generalization of our method to various VAE initialization, we conduct experiments on three commonly used VAEs, including SD-VAE~\cite{stabilityai2025sdvae}, IN-VAE~\cite{leng2025repa} and VA-VAE~\cite{yao2025reconstruction}. The results are shown in~\cref{tab:vae_type}.
As we can see, across all variations, our $\mathcal{L}_{\text{align}}$ can consistently improve final generation performance, which demonstrates that insensitiveness of our method to the VAE initialization.
Moreover, to verify the effectiveness of Send-VAE on larger resolution generation, we also finetune SD-VAE~\cite{stabilityai2025sdvae} using the proposed $\mathcal{L}_{\text{Align}}$ at 512x512 resolution on ImageNet-1k. Then, a Sit-B is trained following the setup in ablation studies. As shown in~\cref{tab:vae_type}, at the 512x512 resolution, the proposed method still significantly improves downstream generative performance, demonstrating the effectiveness of our approach.

\begin{table}[]
\vspace{-20pt}
\centering
\setlength{\tabcolsep}{6pt}
\renewcommand\arraystretch{1.5}
\setlength{\belowcaptionskip}{0pt}
\caption{Ablation on the resolution and initialization of VAE.}
\label{tab:vae_type}
\begin{tabular}{lccccc}
\toprule
VAE Initialization & gFID$\downarrow$ & sFID$\downarrow$ & IS$\uparrow$ & Prec.$\uparrow$ & Rec.$\uparrow$ \\ \midrule
SD-VAE (256 $\times$ 256)             & 21.41 &  5.30 &  65.0  &  0.62  & \textbf{0.63}  \\
\ \ \ \ \ \ +$\mathcal{L}_{\text{align}}$              &   \textbf{11.86}   &  \textbf{5.25} &  \textbf{95.2}  &   \textbf{0.73}    &   0.58   \\ \midrule
SD-VAE (512 $\times$ 512)             & 23.59 &  6.74 &  65.67  &  0.71  & \textbf{0.59}  \\
\ \ \ \ \ \ +$\mathcal{L}_{\text{align}}$              &   \textbf{13.32}   &  \textbf{4.75} &  \textbf{93.15}  &   \textbf{0.78}    &   0.60   \\ \midrule
IN-VAE (256 $\times$ 256)            &  17.43 &  5.93    &  72.7  &  0.64  &  \textbf{0.63} \\
\ \ \ \ \ \ +$\mathcal{L}_{\text{align}}$              &  \textbf{8.25} &  \textbf{4.68} & \textbf{105.2} &  \textbf{0.74}  &  0.60    \\ \midrule
VA-VAE (256 $\times$ 256)            & 11.40 & 6.58 & 93.5 &  0.71  & 0.59 \\
\ \ \ \ \ \ +$\mathcal{L}_{\text{align}}$              &  \textbf{7.57}   & \textbf{5.37} & \textbf{115.3} &   \textbf{0.74}    &   \textbf{0.60}      \\ \bottomrule
\end{tabular}
\vspace{-15pt}
\end{table}

\textbf{Analysis of Reconstruction Performance.}
To explicitly analyze the trade-off between pixel fidelity and generation quality, we evaluate PSNR, SSIM, and LPIPS on ImageNet-1k validation set. As shown in~\cref{tab:reconstruction}, both E2E-VAE and Send-VAE exhibit similar reconstruction profiles (PSNR$\approx$27.6, SSIM$\approx$0.77), which are slightly lower than the Naive VAE. This confirms a community consensus: generation-friendly VAEs prioritize semantic structure over high-frequency pixel noise. Despite similar pixel-level metrics, Send-VAE achieves a better LPIPS (0.101 vs. 0.110) than E2E-VAE, indicating superior perceptual preservation. This marginal drop in PSNR is a necessary cost for semantic disentanglement. Crucially, with comparable reconstruction costs to E2E-VAE, Send-VAE yields superior generation performance and faster convergence.

\begin{table}[]
\vspace{-16pt}
\centering
\setlength{\tabcolsep}{6pt}
\renewcommand\arraystretch{1.5}
\setlength{\belowcaptionskip}{0pt}
\caption{Measurement of the reconstruction performance.}
\label{tab:reconstruction}
\begin{tabular}{cccccc}
\toprule
VAE types & rFID$\downarrow$ & PSNR$\uparrow$                & LPIPS $\downarrow$                     & SSIM$\uparrow$ & gFID$\downarrow$                                       \\ \midrule
Naive VAE & 0.26 & 28.59                                                & 0.089                                                & 0.80 & 17.43                                               \\
VA-VAE    & 0.28 & 27.96                                                & 0.096                                                & 0.79 & 11.40                                               \\
E2E-VAE   & 0.28 & 27.63 & 0.110 & 0.77 & 8.96                                                \\
Send-VAE  & 0.31 & 27.62                                                & 0.101                                                & 0.77 & 7.57 \\ \bottomrule
\end{tabular}
\vspace{-20pt}
\end{table}


\begin{table}[h]
\centering
\setlength{\tabcolsep}{4pt}
\setlength{\belowcaptionskip}{0pt}
\renewcommand\arraystretch{1.5}
\caption{System-level measurement of semantic disentanglement ability of various VAEs. F1 score is adopted for all benchmarks.}
\label{tab:attri}
\begin{tabular}{ccccc}
\toprule
Benchmarks          & IN-VAE & VA-VAE & E2E-VAE                       & Send-VAE \\ \midrule
CelebA     & 0.6222 & 0.6347 & {\color[HTML]{1F2329} 0.6439} & 0.6647   \\
DeepFasion & 0.0786 &  0.1094 &     0.1177                          &  0.1385   \\
AwA        & 0.5567 & 0.5948  &      0.6441              &      0.6623    \\
gFID        & 17.43 & 11.40  &      8.96              &      7.57    \\ \bottomrule
\end{tabular}
\vspace{-10pt}
\end{table}

\subsection{Measurement of semantic disentanglement ability}
To give a system-level measurement of semantic disentanglement capability, we adopt linear probing on attribute prediction benchmarks across distinct domains to measure the semantic disentanglement ability of various VAEs.
Specifically, three attribute prediction benchmarks are used to ensure a comprehensive evaluation, including CelebA~\cite{liu2015faceattributes}, DeepFashion~\cite{liuLQWTcvpr16DeepFashion} and AwA~\cite{lampert2013attribute}.
We conduct linear probing on the flattened latent representation from VAE encoder and show the results in~\cref{tab:attri}.
As we can see, among all benchmarks, the performance of attribute prediction is positively correlated with the down-stream generation performance.
These results strongly support our hypothesis, and making the linear probing on attribute prediction task a suitable metric to evaluate the goodness of a VAE for diffusion.
Meanwhile, we observe that Send-VAE can significantly enhance the semantic disentanglement ability of VAE and achieve superior generation performance.

\section{Conclusion}

In this paper, we investigate the fundamental characteristics that constitute a truly generation-friendly VAE. By revisiting recent representation alignment paradigms, we identify a critical conceptual flaw in prior works: the erroneous assumption that VAEs and LDMs share identical semantic alignment targets. Diverging from this, we empirically demonstrate that while LDMs thrive on abstract semantics, VAEs fundamentally require strong semantic disentanglement to faithfully preserve fine-grained, structured visual elements. This insight is robustly validated by a striking correlation we established between the linear separability of low-level attributes in the latent space and downstream generation quality.
To explicitly cultivate this property, we introduce the Semantic-disentangled VAE (Send-VAE). Rather than forcing a rigid, shallow alignment, Send-VAE leverages a non-linear mapping architecture, which effectively bridges the representation gap by safely absorbing rich contextual guidance from VFMs while actively safeguarding the VAE's native structured semantics. Extensive experiments on ImageNet $256\times256$ confirm the empirical superiority of our approach: Send-VAE not only significantly accelerates the training convergence of flow-based transformers such as SiTs, but also sets a new state-of-the-art generation benchmark, achieving remarkable FID scores of 1.21 and 1.75 with and without classifier-free guidance, respectively.


\clearpage  


%
%
\bibliographystyle{splncs04}
\bibliography{main}
\end{document}